\documentclass[10pt,journal, compsoc]{IEEEtran}
\usepackage{amsmath,amsfonts}
\usepackage[caption=false,font=normalsize,labelfont=sf,textfont=sf]{subfig}
\usepackage{textcomp}
\usepackage{stfloats}
\usepackage{url}
\usepackage{verbatim}
\usepackage{graphicx}
\usepackage[compress]{cite}  
\usepackage{times}                     

\usepackage{tabu}                      
\usepackage{booktabs}                  
\usepackage{lipsum}                    
\usepackage{mwe}                       
\usepackage{multirow}
\usepackage{mathptmx}                  
\usepackage{glossaries}
\usepackage{graphicx}
\usepackage{xspace}                    
\usepackage{svg}
\usepackage{verbatim}

\usepackage{placeins}
\usepackage{adjustbox}
\usepackage{mathptmx}                  
\usepackage{glossaries}
\usepackage{graphicx}
\usepackage{xspace}                    
\usepackage{svg}
\usepackage{verbatim}

\usepackage{placeins}
\usepackage{adjustbox}

\usepackage{enumitem}
\usepackage{rotating}
\usepackage{longtable}
\usepackage{makecell} %
\usepackage{xcolor}

\usepackage{hyperref} 
 
\usepackage{times}                     
\usepackage{mathptmx}                  

\usepackage{placeins}
\usepackage{subcaption}

\usepackage{adjustbox}

\usepackage{booktabs}
\usepackage{tabularx}
\usepackage{multirow}

\usepackage{tikz}
\usetikzlibrary{calc,arrows.meta}

\usepackage{xcolor}

\usepackage{textcomp}
\usepackage{xspace}
\usepackage{verbatim}

\usepackage{enumitem}

\usepackage{setspace}
\usepackage{placeins}
\usepackage{balance}
\usepackage{lineno}

\usepackage{algorithm}
\usepackage{algpseudocode}
\let\Algorithm\algorithm
\renewcommand\algorithm[1][]{\Algorithm[#1]\setstretch{1.5}}

\usepackage{acro}

\usepackage[normalem]{ulem}

\usepackage{tabu}
\usepackage{lipsum}
\usepackage{mwe}

\newcommand{\susheader}[1][6.0]{
\begin{tikzpicture}[baseline=-0.6ex, every node/.style={font=\scriptsize}]
  \def\W{#1}   
  \def\M{0.18} 
  \draw[-{Stealth[length=1.6mm]}] (0,0) -- (\W,0);
  \node[left]  at (0,0) {SD};
  \node[right] at (\W,0) {SA};
  \foreach \i/\n in {0/1,1/2,2/3,3/4,4/5}{
    \pgfmathsetmacro{\x}{\M + \i*(\W-2*\M)/4}
    \draw (\x,0) circle (0.08);
    \node[below=2pt] at (\x,0) {\n};
  }
\end{tikzpicture}%
}

\newcommand{\susscale}[1][6.0]{%
\begin{tikzpicture}[baseline=-0.6ex, every node/.style={font=\scriptsize}]
  \def\W{#1}\def\M{0.18}
  \draw[-{Stealth[length=1.6mm]}] (0,0) -- (\W,0);
  \node[left]  at (0,0) {SD};
  \node[right] at (\W,0) {SA};
  \foreach \i in {0,1,2,3,4}{
    \pgfmathsetmacro{\x}{\M + \i*(\W-2*\M)/4}
    \draw (\x,0) circle (0.08);
  }
\end{tikzpicture}%
}

\usepackage{amsmath,amssymb}
\usepackage{booktabs}
\usepackage{tabularx} 

\usepackage{tikz}
\usetikzlibrary{calc,arrows.meta}

\usepackage{xcolor}

\graphicspath{{figures/}{pictures/}{images/}{./}} 

\usepackage{times}                     

\usepackage{tabu}                      
\usepackage{booktabs}                  
\usepackage{lipsum}                    
\usepackage{mwe}                       

\usepackage{mathptmx}                  
\usepackage{glossaries}
\usepackage{graphicx}
\usepackage{svg} 
\usepackage{xspace}                    

\usepackage{placeins}

\usepackage[capitalise]{cleveref}
\newcommand{\etal}{\textit{et~al.}}

\newcommand{\ie}{i.e., }

\newcommand{\click}{ClickAIXR}

\newcommand{\cxr}{\textit{ClickAIXR}\xspace}

\makeglossaries

\newacronym{llm}{LLM}{Large Language Model} 
\newacronym{adb}{ADB}{Android Debug Bridge}
\newacronym{gguf}{GGUF}{GPT-Generated Unified Format}
\newacronym{ui}{UI}{User Interface}
\newacronym{nn}{NN}{Neural Network}
\newacronym{pp}{PP}{Prompt Processing}
\newacronym{tg}{TG}{Token Generation}
\newacronym{bt}{BT}{Batch Test}
\newacronym{tt}{TT}{Thread Test}
\newacronym{rss}{RSS}{Resident Set Size}

\newacronym{vlm}{VLM}{Vision Language Model} 
\newacronym{xr}{XR}{Extended reality} 
\newacronym{hud}{HUD}{head-up display}
\newacronym{roi}{ROI}{region of interest}
\newacronym{asr}{ASR}{Automatic Speech Recognition}
\newacronym{sus}{SUS}{System Usability Scale}

\begin{document}


\title{ClickAIXR: On-Device Multimodal Vision-Language Interaction with Real-World Objects in Extended Reality}

 

  \vskip -0.365cm
\author{
Dawar Khan,
Alexandre Kouyoumdjian,
Xinyu Liu,
Omar Mena,
Dominik Engel,
and Ivan Viola

 \setcounter{figure}{0}     
  \centering
  \includegraphics[width=0.867\linewidth]{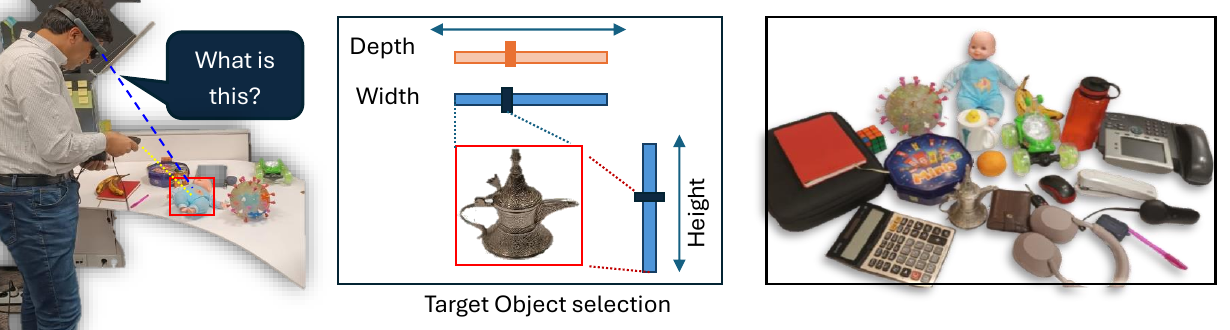}
\captionsetup{margin=0.9985cm}
  \captionof{figure}{Overview of \textbf{ClickAIXR}. \textit{Left:} A user selects a real-world object and queries the on-device VLM (e.g., “What is this?”). \textit{Middle:} The object selection interface, where users adjust a red cropping box via three sliders controlling depth, width, and height. \textit{Right:} Examples of selected objects used in our experiments.}
  \label{fig:teaser}

\IEEEcompsocitemizethanks{ 
\IEEEcompsocthanksitem Dawar Khan,
Alexandre Kouyoumdjian,
Xinyu Liu,
Omar Mena,
Dominik Engel,
and Ivan Viola are with King Abdullah University of Science and Technology (KAUST), Saudi Arabia.\\ E-mail: \{dawar.khan, xinyu.liu,  omar.mena, donggang.jia, alexandre.kouyoumdjian, ivan.viola\}@kaust.edu.sa.  \\
}
\thanks{Manuscript received MM dd, YYYY; revised MM dd, YYYY.\\ (Corresponding author: Dawar Khan.)}}


\markboth{Journal of \LaTeX\ Class Files,~Vol.~14, No.~8, August~2021}%
{Shell \MakeLowercase{\textit{et al.}}: A Sample Article Using IEEEtran.cls for IEEE Journals}

\IEEEpubid{0000--0000/00\$00.00~\copyright~2021 IEEE}

\IEEEtitleabstractindextext{ 
\begin{abstract} 
We present \click{}, a novel on-device framework for multimodal vision-language interaction with objects in extended reality (XR). Unlike prior systems that rely on cloud-based AI (e.g., ChatGPT) or gaze-based selection (e.g., GazePointAR), \click{} integrates an on-device vision-language model (VLM) with a controller-based object selection paradigm, enabling users to precisely click on real-world objects in XR. Once selected, the object image is processed locally by the VLM to answer natural language questions through both text and speech. This object-centered interaction reduces ambiguity inherent in gaze- or voice-only interfaces and improves transparency by performing all inference on-device, addressing concerns around privacy and latency.
We implemented \click{} in the Magic Leap SDK (C API) with ONNX-based local VLM inference. We conducted a user study comparing \click{} with Gemini 2.5 Flash and ChatGPT 5, evaluating usability, trust, and user satisfaction. Results show that latency is moderate and user experience is acceptable.
Our findings demonstrate the potential of click-based object selection combined with on-device AI to advance trustworthy, privacy-preserving XR interactions. The source code and supplementary materials are available at~\href{www.nanovis.org/ClickAIXR.html}{\texttt{nanovis.org/ClickAIXR.html}}.
\end{abstract}
\begin{IEEEkeywords}
Extended Reality, Vision-Language Models (VLMs), Multimodal VLMs, On-Device AI, Gaze Tracking, Privacy, Conversational User Interfaces
\end{IEEEkeywords}
 } 
\maketitle




 \IEEEdisplaynontitleabstractindextext 
 \IEEEpeerreviewmaketitle
\IEEEraisesectionheading{\section{Introduction}}


\gls{xr} environments present unprecedented opportunities for seamless integration of digital information with the physical world. As users navigate real-world scenarios -- whether exploring historical sites, learning complex concepts, or requiring assistance with visual accessibility -- the ability to instantly query and understand their visual environment becomes increasingly valuable. \glspl{vlm} offer compelling capabilities to bridge this gap by enabling natural language interactions with visual content, transforming how users can gain knowledge about what is in their field of view.


Current approaches to integrating \glspl{vlm} into XR applications rely on cloud-based inference, exemplified by systems like \textit{GazePointAR}~\cite{Lee2024}, and\textit{ XaiR}~\cite{Srinidhi2024ISMAR:XaiR}. While these cloud-based solutions leverage powerful computational resources to deliver high-quality responses, they also introduce significant challenges that limit their practical deployment and widespread adoption. Most critically, transmitting potentially sensitive visual data from users' personal environments to proprietary cloud services raises substantial privacy concerns, particularly in educational, workplace, or personal contexts where confidential information is abundant.

Beyond privacy considerations, cloud-based deployments suffer from inherent latency, increased power consumption, recurring subscription costs, and dependence on a stable network connection. Another challenge lies in how to feed the \gls{vlm} with target images or a \gls{roi}. In addition, voice-based interaction is prone to pronoun ambiguity~\cite{tyler1977line} and lacks context awareness. For example, when a user provides an object to the \gls{vlm} and asks ``What is that?'', the pronoun \textit{that} introduces confusion, as the model cannot confirm which object the user is referring to. This issue has recently been addressed by GazePointAR~\cite{Lee2024}, which leverages eye gaze, finger pointing, and a cloud-based VLM to enable ambiguity-free voice assistance and XR interaction. However, two key challenges remain: (i) GazePointAR is cloud-based, so all of the aforementioned limitations persist, and (ii) it relies on YOLOv8~\cite{Jocher_Ultralytics_YOLO_2023} for object segmentation, which introduces additional processing time (reported CV stage: $3.75 \pm 0.23$\,s)~\cite{Lee2024}.

This paper presents a novel approach, \cxr, that addresses these limitations through \emph{local on-device VLM deployment} on XR headsets (Magic Leap~2), leveraging the increased computational power of recent advances in wearable hardware. Our method reduces long, uninterrupted waiting times by bypassing time-consuming image segmentation preprocessing. Instead, it provides users with interactive selection of regions of interest through our Gaze-Locked Clipping Window (GCW). The GCW follows user gaze to position a clipping window on the target object and offers three slider-based GUI controls to adjust the width, height, and depth (distance from the user) of the window (see~\Cref{fig:teaser}, middle), enabling accurate object selection. After selecting a \gls{roi}, users can pose natural language questions about the chosen content. By performing inference entirely on-device, our system eliminates the need to transmit sensitive visual data to proprietary AI providers, thereby ensuring complete privacy preservation.

Despite operating on constrained hardware compared to cloud-based solutions, our approach achieves comparable overall latency with systems like GazePointAR, by removing network delays and pre-processing overhead. Deploying the VLM locally also offers significant advantages in power efficiency, contributing to more sustainable AI applications. We evaluate our method in a within-subjects user study with 12 participants, where we compare \click{} to Google Gemini and OpenAI ChatGPT, and we show that on-device VLM inference can deliver practical and privacy-preserving XR applications with tolerable compromises on user experience.

To summarize, our work makes the following contributions:
\begin{itemize}[leftmargin=*,noitemsep]
  \item  We present (to the best of our knowledge) the first \emph{on-device} multimodal VLM application for XR that supports voice, eye–gaze, text, and image inputs. Running entirely on-device preserves privacy and enables application-specific fine-tuning without sharing data with third parties. The framework is model-agnostic and supports deploying any suitably sized VLM in ONNX format with only minor changes to the inference and tokenizer modules.
  
  \item  We introduce Gaze-Locked Clipping Window (GCW), a gaze-locked, controller-adjustable rectangular window that precisely selects a target object in XR (see Fig.~\ref{fig:teaser}, middle).
  
 \item We conduct a user study and a timing analysis comparing against cloud-based baselines (\ie ChatGPT, Gemini). Our system achieves acceptable usability and practical performance, with mean inference time of \mbox{5.36\,s (Books)–5.48\,s (COCO)} per image and a consistent token-generation speed of \mbox{3.36\,tokens/s}.
\end{itemize}

The remainder of the paper first explores the related work on deploying VLMs on-device and multimodal interaction in XR, followed by our method where we present our solution to the on-device deployment and how we solve for pronoun ambiguity. After that we present and discuss the user study we performed to evaluate our approach and present our conclusions.

\section{Related Work}
\label{sec:related-work}
We review related work along two key research directions: \textbf{(1)} the integration of vision-language models (VLMs) for on-device inference, and \textbf{(2)} the design of human interaction techniques that enable natural, multimodal communication with objects in extended reality (XR). The first line of work emphasizes the importance of privacy, latency, and responsiveness in VLM-powered systems, while the second focuses on resolving referential ambiguity and improving interaction flow through modalities such as gaze, speech, and visual input.

\subsection{On-Device Vision Language Models}
Recent advances in VLMs have significantly improved the ability of systems to understand and generate language grounded in visual context. Models such as \textit{BLIP-2} \cite{li2023blip2} and \textit{LLaVa} \cite{liu2023llava} demonstrate strong visual reasoning. While such models provide robust multimodal capabilities, they are typically designed for server-grade inference. \textit{MobileVLM} \cite{chu2023mobilevlmfaststrong} introduces a family of compact, instruction-following VLMs tailored for mobile CPUs and edge GPUs. Similarly, \textit{MiniGPT-4} \cite{zhu2023minigpt4} and \textit{InstructBLIP}\cite{dai2023instructblip} aim to reduce complexity and improve instruction-following in visual contexts, but still largely depend on training resources and time. 

Recent work on hardware-specific deployment demonstrates the feasibility of running neural networks directly on XR devices. Zaccardi et al. \cite{zaccardi2023device} conducted comprehensive benchmarking of deep learning frameworks on Microsoft HoloLens2, showing that Unity Barracuda significantly outperforms Windows Machine Learning (WinML) for most models, with inference times ranging from milliseconds to seconds depending on model complexity. Hohman et al. \cite{hohman2024model} provide practical insights from 30 industry experts on model compression strategies, highlighting that post-training quantization (fp32 $\rightarrow$ fp16 $\rightarrow$ int8) serves as crucial first step for performance optimization.

TinyVLA~\cite{junjieTinyVLA} introduces compact vision-language-action models that prioritize fast inference and data efficiency, using lightweight backbones and diffusion-based policy heads to enable low-latency deployment without large-scale pretraining. Similarly, PaLM-E~\cite{10.5555/3618408.3618748} explores embodied multimodal reasoning by integrating visual and sensor inputs directly into a \gls{llm}, showing the potential of unified architectures for grounded decision-making, though at a larger scale and higher compute. 

Recent work on \textit{LLM-XR integration} has increasingly focused on the feasibility of deploying \glspl{llm} for \emph{local inference} in spatial computing systems. \textit{LoXR}~\cite{10973004} introduces a benchmark for evaluating the runtime performance, power consumption, and latency of running \glspl{llm} on-device in XR environments. The study compares multiple hardware setups and model configurations, emphasizing the trade-offs between interactivity and model complexity. While LoXR provides valuable insights into the system-level feasibility of on-device inference, it does not address interaction design or user-facing techniques.

Complementary to this work, \emph{AIvaluateXR}~\cite{Khan2025} presents a comprehensive framework for benchmarking \glspl{llm} across multiple XR devices. It benchmarks 17 LLMs on four XR platforms, measuring performance consistency, processing speed, memory usage, and battery consumption across 68 model--device pairs. The framework further conducts a Pareto analysis to identify optimal device--model configurations, and compares on-device inference against client--server and cloud-based setups. Although \emph{AlvaluateXR} targets LLMs (not VLMs), it includes experiments on two XR datasets; the authors conclude that both LLMs and VLMs are feasible for on-device XR applications, with accuracy improvable via fine-tuning on XR data and efficiency gain achievable through model compression and quantization.

\subsection{Multimodal Interaction in XR and Voice Assistants}

A significant challenge in XR AI agents is resolving referential ambiguity in user queries. The foundations for multimodal pronoun disambiguation were established in earlier work. Lee et al. \cite{lee2021whats} pioneered the \textit{TouchVA} system, which combined touch and voice for demonstrative pronoun disambiguation, establishing foundations for spatial reference resolution in mobile contexts.

The \textit{GazePointAR} system~\cite{Lee2024} addresses pronoun ambiguity by combining static eye gaze, pointing gestures, computer vision techniques, and a cloud-based \gls{llm}  to disambiguate pronouns in real time. \textit{Walkie-Talkie} \cite{lee2025walkietalkie} advances beyond static gaze capture to dynamic gaze patterns combined with LLMs and Vision-Language Models for query disambiguation, representing an evolution from GazePointAR's single-moment gaze capture to continuous tracking.
While this enables more natural multimodal interaction, the system heavily relies on cloud processing, introducing concerns around latency, privacy, and transparency. Additionally, its reliance on gaze as the primary selection modality can lead to user fatigue and reduced accuracy, particularly in crowded or dynamic environments.

Torre \etal~\cite{de2024llmr} propose \emph{LLMR}, which leverages \glspl{llm} for real-time creation and modification of interactive mixed-reality content, enabling tasks such as generating new assets or editing existing elements directly on VR/AR devices. \textit{XaiR}~\cite{Srinidhi2024ISMAR:XaiR} integrates multimodal LLMs (MLLMs) with XR using a client--server architecture: computationally intensive MLLM inference is offloaded to a server while spatial context is handled locally on the headset. The system supports real-time multimodal input; however, end-to-end latency, stemming from both network delays and LLM inference time remains a practical challenge.

Expanding the boundaries of multimodal input, the \textit{GesPrompt} system~\cite{Hu2025} augments spoken interaction in VR with \emph{co-speech gestures}, allowing users to enrich their language prompts with spatio-temporal gestures. 
This approach mirrors natural human communication, helping to reduce the cognitive burden of crafting detailed textual descriptions. 
However, like GazePointAR, GesPrompt can still suffer from ambiguity when multiple objects are present in the scene, as it lacks explicit object selection mechanisms. 

Complementing prior systems, Wang et al.\cite{wangSpatial2025} provide a comprehensive review of recent multimodal interaction techniques in XR, highlighting the increased use of gaze, speech, and gesture combinations to address user fatigue and ambiguity in selection. While large models, such as PaLM-E, offer robust referential understanding, their scale remains a barrier to on-device use.

\begin{figure*}[!htbp]
    \centering 
    \includegraphics[width=0.998\linewidth]{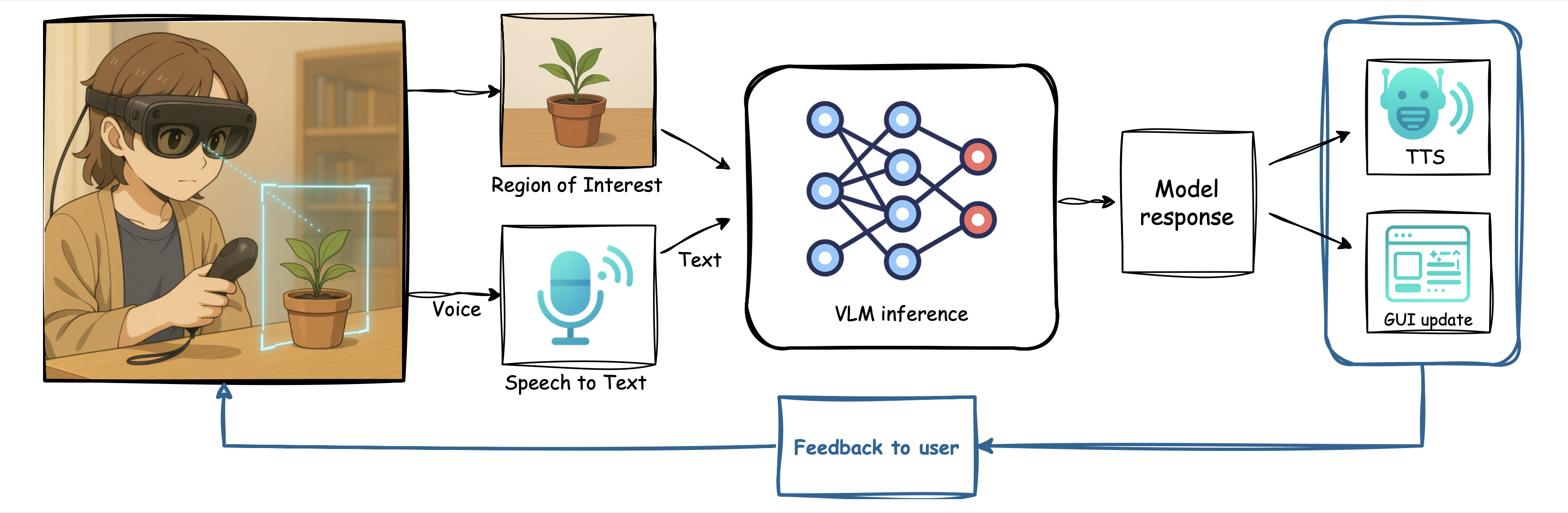} %
\caption{Overview of the \cxr\ pipeline. Users choose between (i) \emph{dwell mode}, where a fixed-size GCW follows gaze and, after a brief dwell, auto-captures an ROI for image captioning or a spoken/text query; or (ii) \emph{GCW select-and-ask}, where the user places the border-only GCW on the target, adjusts width/height/depth with the controller, and confirms with a trigger. After confirmation, a microphone icon appears; the spoken question is converted to text via on-device ASR, then fused with the cropped image and processed by the on-device \gls{vlm} (encoder–decoder, tokenizer). The answer is returned to the XR UI as text and to the user as audio via TTS on ML2.}
    \label{fig:pipeline}
\end{figure*}

\section{Research Method and Materials}
\label{sec:method}
To advance natural, low-ambiguity XR interaction while preserving privacy and keeping latency predictable, we designed and built \cxr: a fully \emph{on-device} multimodal assistant that couples a local \gls{vlm} with gaze-locked object selection and on-device speech I/O. Co-locating perception and inference on the headset removes network dependencies and API costs, reduces tail latency, and enables trustworthy, object-grounded exchanges (e.g., disambiguating pronouns by selection rather than intent inference).

We implemented \cxr using the Magic Leap~2 (ML2) device and its MLSDK C API. The \gls{vlm} runs entirely on-device via ONNX Runtime~\cite{onnxruntime}; \gls{asr} is provided by an on-device Vosk model~\cite{vosk}, and answers are presented both as on-screen text and via on-device TTS. The \gls{vlm} and \gls{asr} models are sideloaded into the app’s files directory on ML2.
\Cref{fig:pipeline} presents the overall pipeline of the \cxr. At launch, a main GUI exposes two usage modes: \emph{dwell auto-capture} (a fixed-size GCW that follows gaze and captures after a short dwell) and \emph{GCW select-and-ask} (the user sizes the border-only GCW with the controller and confirms with the trigger). After capture, the cropped ROI and the spoken/text query are processed entirely on-device by the encoder–decoder and tokenizer; the answer is returned to the XR UI and optionally read aloud.

\subsection{On-Device Multimodal VLM}
We deploy a lightweight \gls{vlm} entirely on-device on Magic Leap~2 (ML2) using the MLSDK C API and ONNX Runtime~\cite{onnxruntime}. The model is the ViT–GPT-2 image-captioning checkpoint~\cite{nlpconnect2023_vitgpt2_captioning}, a VisionEncoderDecoder architecture coupling a ViT-Base image encoder~\cite{dosovitskiy2021vit} with a GPT-2 language decoder~\cite{radford2019gpt2}. We export the checkpoint to ONNX with Hugging Face Optimum (\texttt{task=image-to-text}, opset~17), producing separate encoder and decoder-with-past (KV-cache) graphs~\cite{optimum}. At runtime, these graphs are executed locally, with all pre-/post-processing on-device (resize to $224{\times}224$ and per-channel normalization as specified in the model card). This configuration enables fully local, privacy-preserving inference without network connectivity. All experiments use the same exported weights and tokenizer, and the code path follows the Transformers stack for parity with the reference implementation.

Our MLSDK C-API pipeline crops the GCW-selected region, captures the user's spoken query, runs encoder–decoder generation (greedy) on-device, and presents the response in the XR UI (with optional TTS). Because preprocessing, inference, and decoding all run locally, the system avoids network variance, thereby reducing latency and preserving privacy. Although the current model was originally trained for captioning rather than instruction following, it reliably handles simple description-oriented queries (e.g., object identity, color, and coarse spatial relations) that suffice for common XR interactions such as ``What is this?''. The framework is \emph{model-agnostic}: any ONNX-exportable VLM that fits the device memory budget can be swapped in by replacing the graphs and tokenizer files with minimal code changes, including instruction-tuned alternatives (e.g., BLIP-2, LLaVA, Qwen-VL) or quantized/distilled variants for tighter resource budgets. While in this paper we use the checkpoint \emph{as is} (no additional training), the same pipeline supports training or fine-tuning a suitably sized multimodal VLM on in-house datasets for application-specific XR tasks and then deploying it on-device for cost-effective, privacy-preserving inference, \ie capabilities that are often difficult to realize with cloud-hosted APIs. 

\subsection{Gaze-Locked Clipping Window (GCW)}
\label{sec:gcw}
In contrast to GazePointAR~\cite{Lee2024}, which relies on YOLOv8~\cite{Jocher_Ultralytics_YOLO_2023} for additional image processing (reported CV stage: $3.75\pm0.23$\,s), our system performs \emph{segmentation-free} target selection via a gaze-locked clipping window (GCW). We render a thin, \emph{border-only} rectangle on a fully transparent, \gls{hud}. The rectangle continuously follows the eye–gaze intersection with the HUD and can be resized with controller inputs by adjusting slider for width and height, providing fast, predictable region-of-interest (ROI) selection without any pixel-wise inference.

The user positions the GCW over the target object (gaze-aligned), optionally adjusts \emph{width/height} and depth \ie the \emph{HUD distance} (a comfort control for the plane’s placement), and confirms selection with a single controller trigger (see~\Cref{fig:teaser}). Upon trigger, we crop the ROI and send it, together with the user’s voice query (from \gls{asr}), to the on-device \gls{vlm}. The VLM processes the image and text and presents the answer in the XR UI as well as via TTS. This mode provides explicit, low-ambiguity selection with minimal risk of accidental activation.

Let $(H_x,H_y)$ denote the HUD half-sizes (m) and let $a=\tfrac{W}{H}$ be the camera aspect ratio. We fit the image rectangle inside the HUD while preserving $a$, with scaling factor $s$:
\begin{equation}
s=\min\!\left(H_y,\ \frac{H_x}{a}\right),\qquad (S_x,S_y)=(sa,\ s).
\end{equation}
A gaze hit on the HUD with local coordinates $(x,y)\in[-S_x,S_x]\times[-S_y,S_y]$ maps to image-normalized coordinates
\begin{equation}
u=\tfrac{1}{2}\!\left(1+\frac{x}{S_x}\right),\qquad
v=\tfrac{1}{2}\!\left(1-\frac{y}{S_y}\right),
\end{equation}
which define the window center $c=(u,v)\in[0,1]^2$. The window size $s_n=(w_n,h_n)$ is maintained in image-normalized units and clamped to remain inside the image ($\tfrac{w_n}{2},\tfrac{h_n}{2}\le 0.49$).

To guarantee pixel-accurate correspondence between the displayed rectangle and the saved crop, we \emph{latch} $(c,s_n)$ at shutter time $t^\star$ and convert to pixel bounds:
\begin{align}
x_0 &= \left\lfloor W\!\left(c_x-\tfrac{w_n}{2}\right)\right\rfloor,\quad
x_1 = \left\lceil  W\!\left(c_x+\tfrac{w_n}{2}\right)\right\rceil,\\
y_0 &= \left\lfloor H\!\left(c_y-\tfrac{h_n}{2}\right)\right\rfloor,\quad
y_1 = \left\lceil  H\!\left(c_y+\tfrac{h_n}{2}\right)\right\rceil.
\end{align}
We then copy $(x_0\!:\!x_1,\ y_0\!:\!y_1)$ from the camera frame and store it as a JPEG. To avoid jitter, a brief fixation gate is applied, requiring the gaze to remain within time and angular thresholds, thereby stabilizing the sample prior to capture.%

In this way, the GCW yields (i) segmentation-free ROI selection, (ii) content-independent overhead dominated by a rectangular memory copy, (iii) zero network dependency, and (iv) exact visual–pixel alignment because the HUD-aligned window state is latched at $t^\star$.

\subsection{GCW Auto-Capture Mode (Fixed Size)}
Beyond manual GCW operation, we provide a simple gaze-driven dwell mode. The user presets the GCW \emph{width} and \emph{height} (fixed size) and the HUD distance in the main GUI; the rectangle’s center then continuously follows the user’s gaze. When the gaze remains within the window for a dwell interval $\tau_{\text{dwell}}$ (configurable), the system latches the current GCW state, crops the corresponding image region, and immediately runs the on-device \gls{vlm}. By default, we issue a short automatic prompt (e.g., “What is in the image?”), unless the user speaks a specific question, in which case that query is used.

 \subsection{Speech I/O Module}
\label{subsec:speech-io}
We use the Vosk on-device \gls{asr} toolkit~\cite{vosk} to provide fully offline speech recognition. Although Vosk supports many languages, in this work we use the English model, bundled with the application and loaded from app-private storage at launch; no network connectivity is required. Speech output is produced with the platform’s built-in \gls{asr}.

\textbf{Streaming \gls{asr}.}
We process audio in streaming mode at 16\,kHz and surface both partial and final transcripts. After each image capture we adopt a \emph{listen-until-silence} policy: the utterance is committed when a short silence grace interval elapses or a maximum timeout is reached, keeping the query field responsive while the user speaks and reducing premature submissions. The main GUI lets users choose either voice-only direct interaction or an editable mode in which a live text field updates in real time; before forwarding to the VLM, users may revise the text via a virtual keyboard or select \emph{Clear} to re-record.

\paragraph{Embedded TTS.}
For responses, we synthesize speech locally using the device locale and standard speaking parameters. We request transient audio focus for playback and release it on completion; recognition is paused during TTS to avoid acoustic feedback. All audio processing remains on-device, which yields predictable latency under poor connectivity and simplifies privacy for mixed-reality use.

\section{Experiments and Results}
\label{sec:results}
This section evaluates \cxr\ along two axes: (i) system latency for on-device VLM inference on public image datasets, and (ii) user experience via a user study.   

\begin{figure*}[t]
    \centering
    \includegraphics[width=0.985\linewidth]{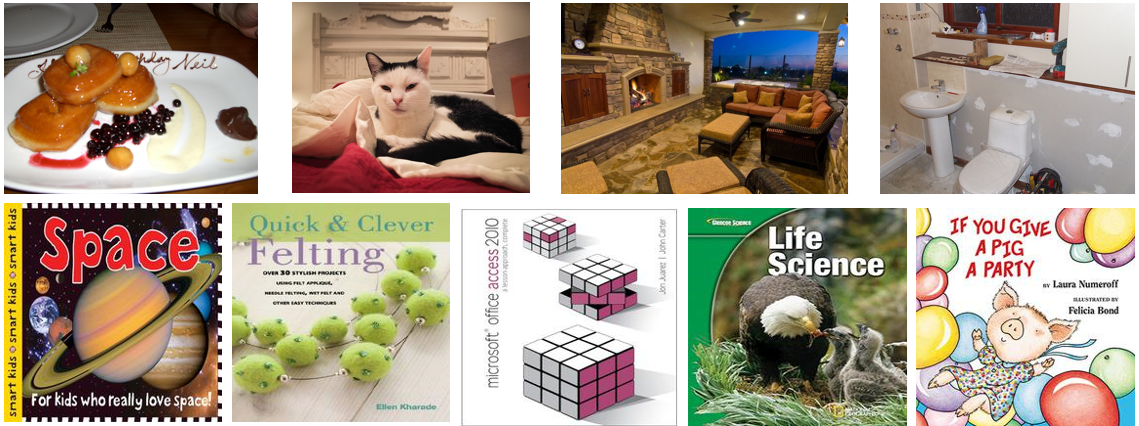}
  \caption{Examples of images used for the latency test: top row from COCO~\cite{LinCoco2014}, bottom row from the Book Covers dataset~\cite{iwana2016judging}.}
    \label{fig:CocoBooks}
\end{figure*}

\begin{table*}
\small
\centering
\setlength{\tabcolsep}{6pt} 
\caption{\gls{sus}~\cite{brooke1996sus} questionnaire items with 5-point Likert responses (1=Strongly Disagree, 5=Strongly Agree). }

\begin{tabularx}{\linewidth}{@{}cXc@{}}
\toprule
\# & \multicolumn{1}{c}{\gls{sus} item} & \multicolumn{1}{c}{Scale: 1 = Strongly Disagree (SD) — 5 = Strongly Agree (SA)} \\
\cmidrule(lr){3-3}
 & & \susheader[6.2] \\  
\midrule
1  & I think that I would like to use this system frequently. & \susscale[6.2] \\
2  & I found the system unnecessarily complex.               & \susscale[6.2] \\
3  & I thought the system was easy to use.                    & \susscale[6.2] \\
4  & I think that I would need the support of a technical person to be able to use this system. & \susscale[6.2] \\
5  & I found the various functions in this system were well integrated. & \susscale[6.2] \\
6  & I thought there was too much inconsistency in this system. & \susscale[6.2] \\
7  & I would imagine that most people would learn to use this system very quickly. & \susscale[6.2] \\
8  & I found the system very cumbersome to use.               & \susscale[6.2] \\
9  & I felt very confident using the system.                  & \susscale[6.2] \\
10 & I needed to learn a lot of things before I could get going with this system. & \susscale[6.2] \\
\bottomrule
\end{tabularx}
\label{tab:sus_items}
\end{table*}

\begin{figure*}
    \centering
    \includegraphics[width=0.9985\linewidth]{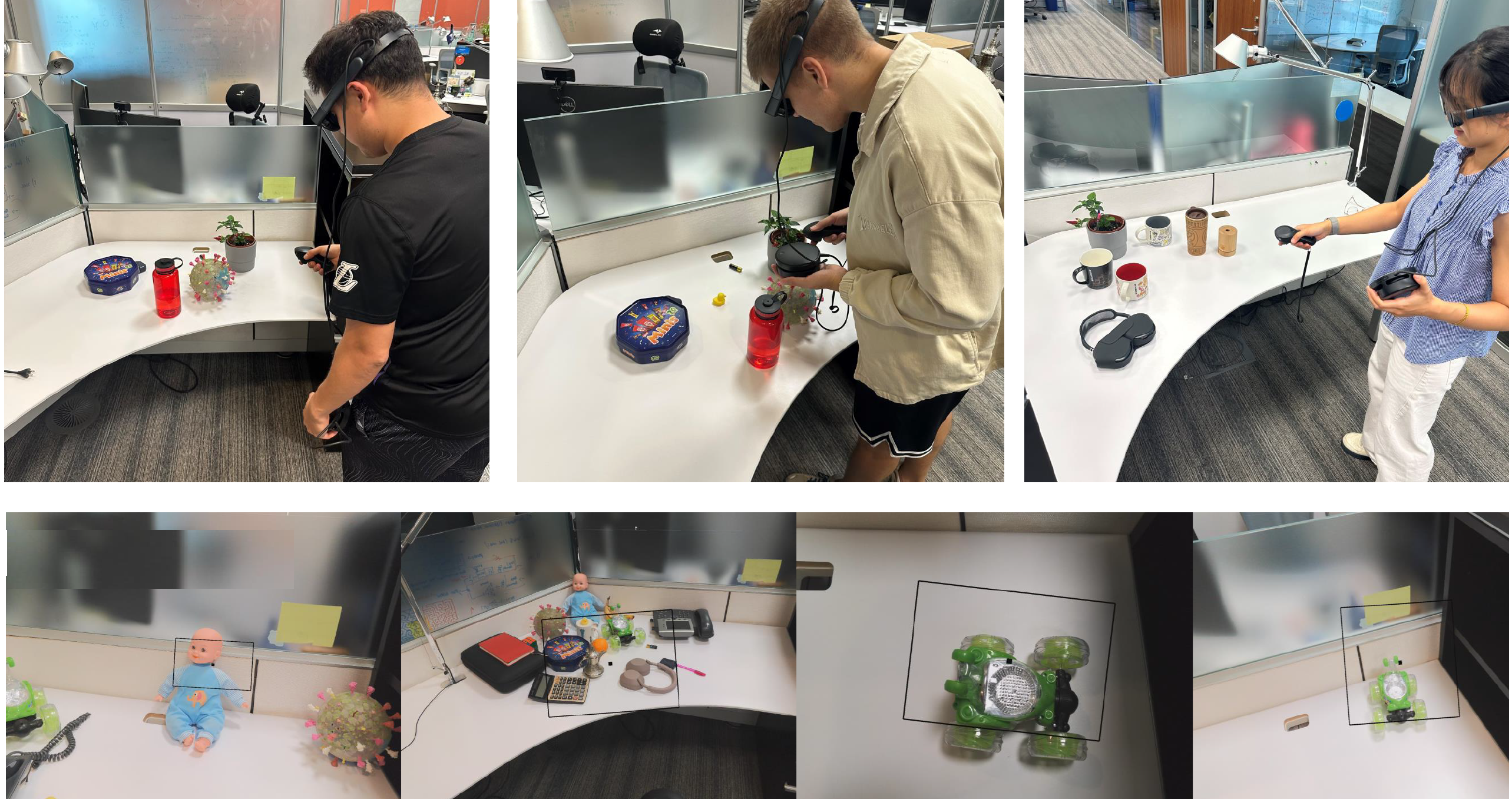}
  \caption{User study overview and in-headset views on Magic Leap~2. 
Top: participants interacting with the object table using \cxr{}. 
Bottom: representative object layouts and direct in-headset views; the border-only rectangle is the gaze-locked clipping window (GCW), which participants are positioning and resizing before capture.}
    \label{fig:exp}
\end{figure*}
 \subsection{Latency Measurements}
 \label{sec:latency}
We evaluated our system on two datasets: 100 images from the Book Covers dataset~\cite{iwana2016judging}, which contains fixed-size images of $224 \times 224$ pixels, and 100 indoor scene images from the COCO dataset~\cite{LinCoco2014}, which contain variable-sized images. These are not cropped objects but full images (see~\Cref{fig:CocoBooks}). 
For each image, we posed the simple question ``What is in the image?'' and recorded (i) the inference time, which includes image encoding and the generation of a short one-line answer, and (ii) the token generation (TG) speed. 
Model loading time was excluded from these measurements --- it was tested separately over 10 runs, and ranged between 3.24 and 3.59 seconds, with a mean of 3.51 seconds. 
The results are summarized in~\Cref{tab:vlm_summary}.

\begin{table}[t]
\caption{On-device VLM performance on the Book Covers dataset~\cite{iwana2016judging} and the COCO dataset~\cite{LinCoco2014}. Inference times are reported in seconds; TG speed is tokens per second.}
\centering
\small
\resizebox{\linewidth}{!}{
\begin{tabular}{lccccc}
\toprule
\textbf{Metric} & \textbf{Mean} & \textbf{Std} & \textbf{Median} & \textbf{Min} & \textbf{Max} \\
\midrule
Inference Time (Books) [s] & 5.36 & 0.80 & 5.03 & 4.36 & 7.20 \\
Inference Time (COCO) [s]  & 5.48 & 1.02 & 5.29 & 4.37 & 8.96 \\
TG Speed (Books) [tokens/s] & 3.36 & 0.06 & 3.38 & 3.21 & 3.45 \\
TG Speed (COCO) [tokens/s]  & 3.36 & 0.04 & 3.37 & 3.22 & 3.44 \\
\bottomrule
\end{tabular}
}
\label{tab:vlm_summary}
\end{table}


 \subsection{User Study}
To evaluate the user experience of \click{} and validate the feasibility of on-device VLM inference, we conducted a study with 12 participants (seven male and five female). We instructed them to request captioning of real-world objects from the system, as shown in \Cref{fig:study_objects}. We hypothesized that users would find the experience satisfactory, despite the use of a much smaller model than the foundational ones run by common applications such as ChatGPT or Gemini.

    \subsubsection{Design}
We performed a within-subjects experiment with a single independent variable: the captioning method used: OpenAI ChatGPT 5, Google Gemini 2.5 Flash, or \cxr.  The first two methods were evaluated on an Android smartphone (Redmi Note~11S), reflecting a common usage scenario. 
\cxr ran on Magic Leap~2 entirely offline; the cloud baselines (Gemini and ChatGPT) used their vision-enabled live-camera mode over a high-bandwidth UniFi Wi\mbox{-}Fi network (92/160\,Mbps down/up; 4/493\,ms unloaded/loaded latency), measured with fast.com\footnote{Mean of 3 runs; \url{https://fast.com}}.

\begin{figure}
    \centering
    \includegraphics[width=0.85\linewidth]{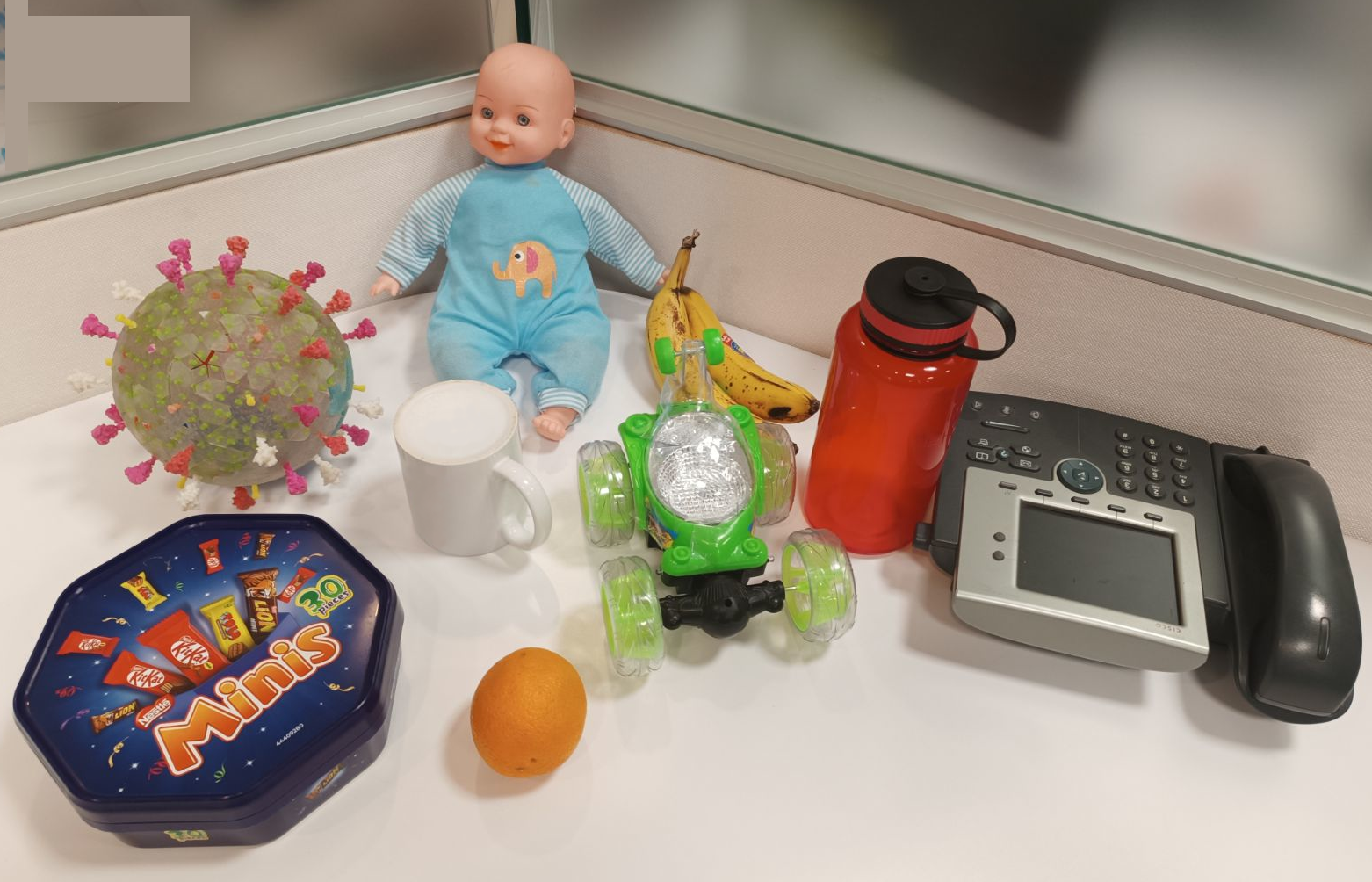}
    \caption{Some of the objects we placed in the room used for the study. Their close proximity to one another often leads to overlap once photographed from a particular angle, which requires disambiguation.}
    \label{fig:study_objects}
\end{figure}

The study took place in a large indoor area with moderate levels of ambient noise, similar to real-world use. \Cref{fig:exp} shows multiple pictures of the participants and the objects they are capturing.  For each method, they were asked to look at the various objects in the room, including, but not limited to, the ones we deliberately placed there, as shown in \Cref{fig:study_objects}.

For each method, the evaluation lasted around 10 minutes. The participants freely wandered around the study area, inquiring about 15 objects for each method.

After completing the evaluation itself, the participants completed one \gls{sus}~\cite{brooke1996sus} questionnaire (see~\Cref{tab:sus_items}) for each method. After the evaluation, our participants also completed a shorter questionnaire with the following, task-specific questions and statements (they would express their agreement with each statement on a 5-point Likert scale):
\begin{enumerate} [leftmargin=*,noitemsep]
    \item How often do you use Augmented Reality?
    \item I could reliably select the intended object even when it was surrounded by other objects.
    \item I often selected the wrong object when multiple objects were present.
\end{enumerate}

Finally, our participants were asked to rank the methods from 1 (best) to 3 (worst). These rankings  provide a preference measure complementary to the \gls{sus} scores.  Ranking was collected immediately after the questionnaires using a single on-screen form; participants assigned distinct ranks 1–3 (no ties permitted). To minimize affiliation bias, the prompt explicitly stated that none of the three systems (Gemini, ChatGPT, \click{}) were ours.

\begin{table}[!htbp]
\centering
\caption{Summary of \gls{sus} results (0--100). Mean $\pm$ SD, median, quartiles, SEM, and $\pm$95\% CI for each method ($n=12$).}
\label{tab:ress12}
\resizebox{1.02\linewidth}{!}{
\begin{tabular}{lcccccccc}
\toprule
Method & $n$ & Mean & SD & Median & Q25 & Q75 & SEM & $\pm$95\% CI \\
\midrule
Gemini    & 12 & 81.88 & 11.24 & 83.75 & 73.75 & 90.63 & 3.24 & 6.36 \\
ChatGPT   & 12 & 76.67 & 15.79 & 76.25 & 69.38 & 90.00 & 4.56 & 8.93 \\
ClickAIXR & 12 & 60.00 & 17.06 & 61.25 & 50.00 & 70.63 & 4.92 & 9.65 \\
\bottomrule
\end{tabular}
}
\end{table}

\begin{figure}
    \centering
    
    \includegraphics[width=\linewidth]{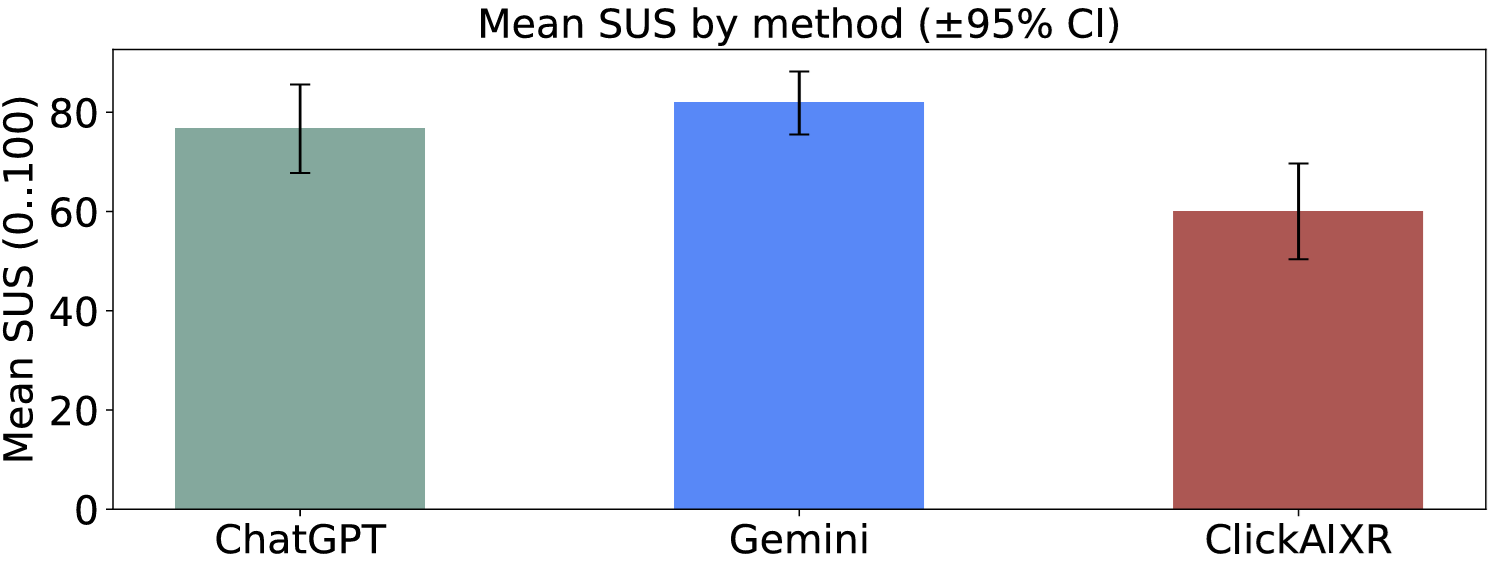}
    \caption{Mean \gls{sus} scores (0--100) for ChatGPT, Gemini, and ClickAIXR.
Bars show mean values; error bars indicate $\pm$95\% confidence intervals (CI).
Results: Gemini = $81.9 \pm 11.2$ (SD), CI $\pm 6.36$;
ChatGPT = $76.7 \pm 15.8$ (SD), CI $\pm 8.93$;
ClickAIXR = $60.0 \pm 17.1$ (SD), CI $\pm 9.65$.}
    \label{fig:sus_all_ci}
\end{figure}


\subsubsection{Results}

We report aggregate \gls{sus} results in \Cref{tab:ress12} and visualize them in \Cref{fig:sus_all_ci}. Results show that the participants found Gemini and ChatGPT to provide greater usability than \click{}. This may be due to the more polished nature of these highly successful commercial applications, but also to their greater ease of deployment, since they were evaluated on a smartphone and did not require the participant to wear an augmented reality device. For such a simple task, the AR device may be seen as more of a constraint than an asset.

\begin{figure}[b]
    \centering
    
    \includegraphics[width=\linewidth]{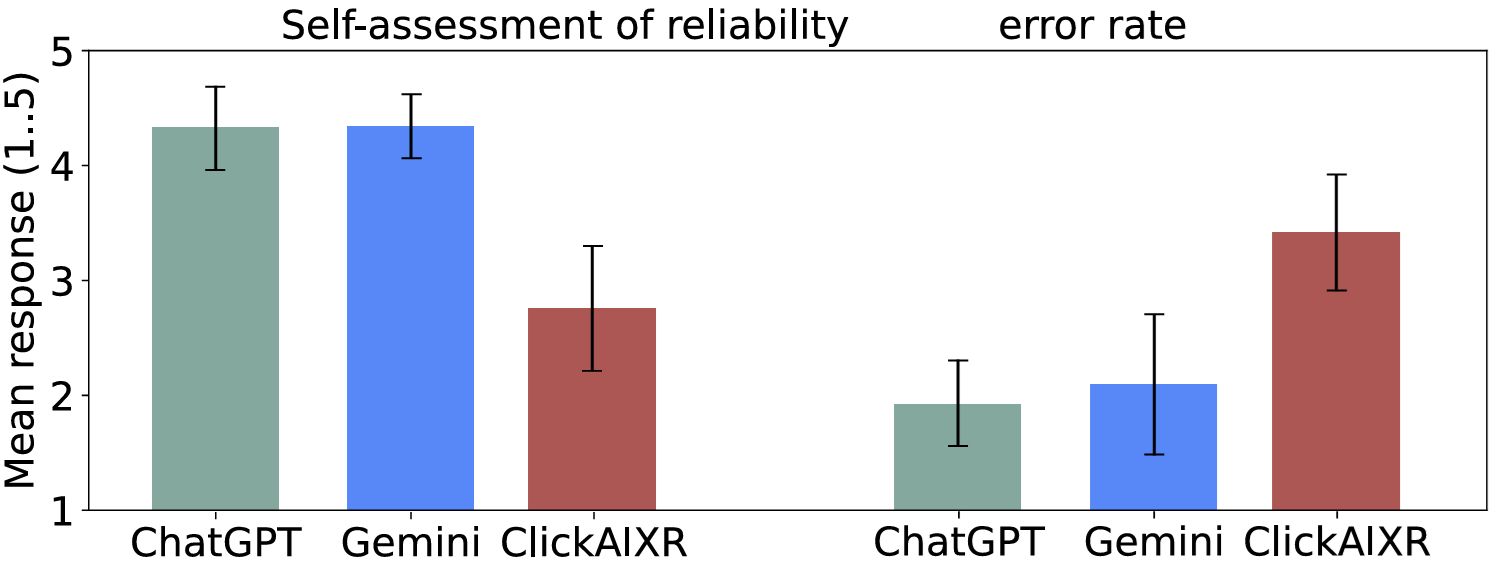}
    \caption{Mean self-assessment ratings (1--5 Likert scale) for reliability (left) and error rate (right) across ChatGPT, Gemini, and \click{}. 
Bars show mean responses; error bars indicate $\pm$95\% confidence intervals (CI). Participants rated ChatGPT and Gemini higher on perceived reliability, while \click{} received lower reliability scores and higher perceived error rates.}
\label{fig:question_mean_ci_reliably}
\end{figure}

\begin{figure} [t]
    \centering
    
    \includegraphics[width=\linewidth]{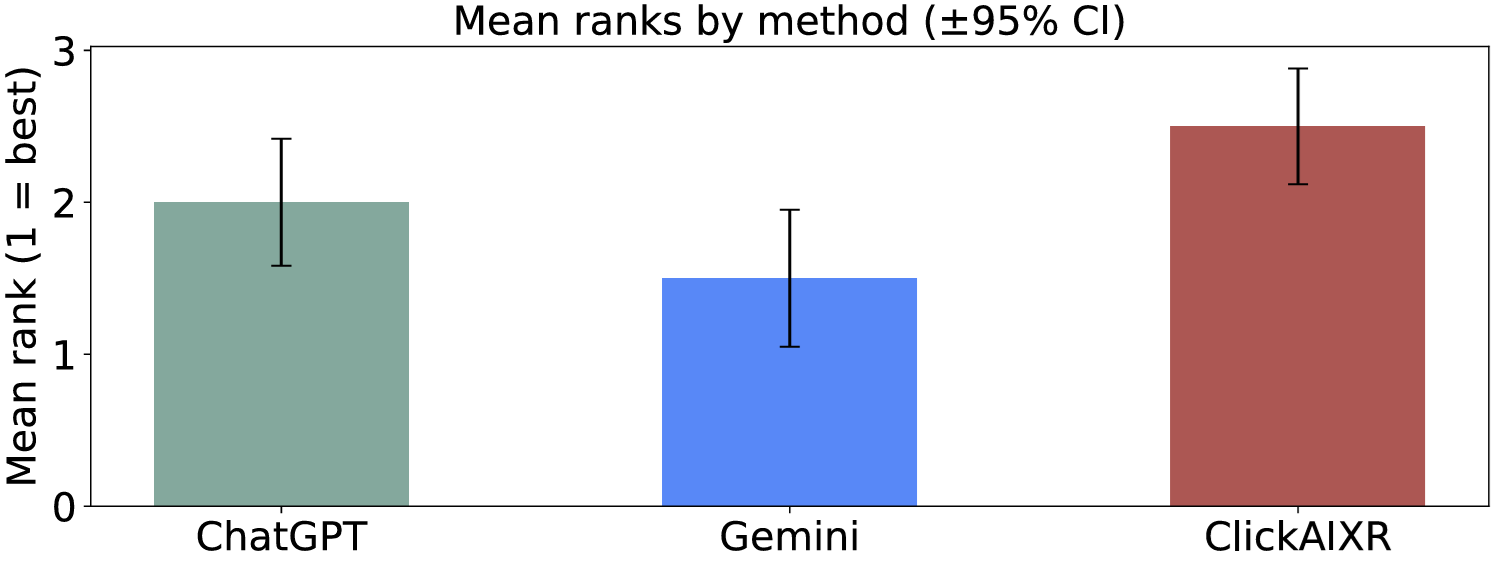}
    \caption{Mean ranks by method (1 = best), with error bars showing $\pm$95\% confidence intervals (CI). Gemini achieved the best average rank, followed by ChatGPT, while \click{} was ranked lowest on average.}

    \label{fig:method_mean_rank_ci}
\end{figure}

The results of the short, task-specific questionnaire are reported in \Cref{fig:question_mean_ci_reliably}. \click{} was found to be significantly less reliable than the other two methods.\\
Finally, the mean rankings assigned by our participants are reported on \Cref{fig:method_mean_rank_ci}. Gemini leads, though its 95\%{} confidence interval significantly overlaps with ChatGPT's. \click{}, however, is behind.\\
\noindent{}\textbf{Interpreting \gls{sus} and positioning:}
Our aggregate \gls{sus} for \click{} was $60.0\pm17.1$, which is comparable to the \gls{sus} reported for \emph{GazePointAR} ($62.1\pm20.0$) in a similar AR context~\cite{Lee2024}. In the \gls{sus} literature, a score of 68 is commonly cited as the overall average benchmark; values in the low 60s fall into the ``marginal/OK'' or ``D–C'' band on adjective/curved grading interpretations~\cite{lewis2018item}. Despite this baseline, our approach provides clear system-level advantages for AR: it is fully on-device, avoids segmentation (and thus pixel-wise inference) by using a gaze-locked clipping window, removes network dependencies, and yields exact visual–pixel alignment via latching, which are desirable traits for latency- and privacy-sensitive XR use~(see Sec.~\ref{sec:gcw}).\\
In terms of efficiency and system-level performance, our approach operates as a standalone solution with fully on-device AI, which is particularly advantageous for XR applications. GazePointAR employs a multi-stage pipeline comprising image capture (2.27\,s), segmentation via YOLOv8 (3.75 $\pm$ 0.23\,s), and cloud-based VLM inference (1.87\,s), resulting in a total reported latency of approximately 7.51\,s~\cite{Lee2024}. 

In contrast, our latency measurements (see \Cref{tab:vlm_summary,sec:latency}) focus on on-device VLM inference (5.36--5.48\,s) and exclude image acquisition time. While these values are not directly comparable due to differences in experimental setup, our method eliminates both the segmentation stage and network-dependent inference, consolidating the pipeline into a single on-device processing step.  


\noindent{}\textbf{Threats to validity:}
First, the comparison used strong, familiar smartphone baselines, which can depress relative \gls{sus} for a novel head-worn interface. Second, \gls{sus} varies by product category and user familiarity~\cite{Bangor2009,Lewis2018}; early AR prototypes often score below web/mobile baselines. Future work will evaluate \click{} on XR-native tasks (hands-busy, heads-up, in-situ object reference) where its on-device, segmentation-free design should matter more. 

While the smartphone baselines are simpler and more familiar, \cxr on Magic Leap~2 enables \emph{heads-up, hands-free, low-profile} capture: gaze-aligned cropping does not require raising or aiming a handheld camera, which typically signals capture. This unobtrusive interaction benefits blind/low-vision assistance and hands-busy settings (e.g., industrial/clinical). We also surface capture events in the UI and keep inference fully on-device to support social acceptability and privacy.

\section{Discussion}

\label{sec:discussion}
To our knowledge, \cxr is the first system to demonstrate fully \emph{offline} VLM interaction on the Magic Leap~2 (ML2) headset. Running entirely on-device offers the following practical benefits: (i) privacy (no visual data leaves the headset), (ii) predictable latency without network variance or API quotas, (iii) operational sustainability (no subscription fees and reduced dependence on energy- and cost-intensive cloud compute), and (iv) support for application-specific fine-tuning on local data, which is often infeasible with proprietary cloud models. These properties make \cxr suitable for scenarios with limited or restricted connectivity (e.g., remote sites or institutionally regulated environments). Most prior systems depend on cloud-based AI; \emph{AlvaluateXR}~\cite{Khan2025} is a notable exception, but it deploys only \gls{llm}s, not \gls{vlm}s.

The proposed Gaze-Locked Clipping Window (GCW) enables \emph{segmentation-free} object selection by cropping the user-specified \gls{roi} directly from the camera frame. This design reduces ambiguity in voice interactions—particularly pronoun ambiguity~\cite{tyler1977line}, because users explicitly select \emph{which} object the VLM should consider before speaking (e.g., isolating only the nose rather than the entire face). In contrast, GazePointAR~\cite{Lee2024} employs YOLOv8-based processing, adding a reported computer-vision stage of \(3.75\pm0.23\,\text{s}\)~\cite{Jocher_Ultralytics_YOLO_2023}. \cxr bypasses this stage entirely by replacing segmentation with user-driven cropping. While ROI selection itself takes a moment, that time is spent in purposeful interaction rather than passive waiting, and users typically do not perceive it as additional latency.

Beyond their time and memory costs, segmentation-based pipelines can still leave \emph{referential} (pronoun) ambiguity unresolved, especially in fine-grained cases, thus exacerbating pronoun ambiguity~\cite{tyler1977line}. For example, when a user points near a person’s nose, a detector may return a mask for the entire face or head, leaving it unclear whether the query targets the nose or the face (as may occur in GazePointAR~\cite{Lee2024}). In contrast, our GCW provides pixel-accurate, user-driven cropping: the selected region is exactly what the VLM receives, improving transparency, controllability, and user trust.

Although our current checkpoint is trained for captioning (not instruction following), it already handles simple description-oriented queries (identity, color, coarse relations). The framework itself is \emph{model-agnostic}: any ONNX-exportable multimodal model that fits the device budget can be swapped in with minimal changes (model directory and tokenizer), including instruction-tuned, quantized, or distilled variants.

The latency measurements reported in \Cref{sec:latency} and \Cref{tab:vlm_summary} show that even when running on the CPU of an XR device, a VLM can provide information about an image within a few seconds. While this is higher than the one-second recommendation to keep the user's flow of thought uninterrupted, it remains below the 10-second limit after which the user's focus would be lost~\cite{nielsen1994usability}. Given that the VLM is not running on the GPU and that the Magic Leap 2 is almost three years old, this result is very encouraging for the future of direct VLM execution on XR devices, especially as dedicated tensor processing units become more common on XR devices, and as VRAM capacities increase.

The reliability results reported in \Cref{fig:question_mean_ci_reliably}  mirror each other and show that the cropping mechanism used in \click{} still has room for improvement. This likely affected the SUS results to a large extent. Future work should focus on addressing this weakness, potentially closing the gap with server-based foundational models.

Our SUS scores were modest. We do not take this to imply that the interface is unusable. Rather, two factors likely depressed ratings. First, participants appear to have \emph{anchored} their judgments on powerful cloud baselines (ChatGPT and Gemini), which set a high reference point for perceived quality. Second, our tasks emphasized relatively simple object–description interactions; such tasks can produce ceiling effects for cloud systems while underrepresenting scenarios where \cxr’s on-device, segmentation-free workflow offers clearer advantages (e.g., privacy-critical or connectivity-limited use, or precise ROI selection).
In post-study comments, several participants noted that if the comparison were a still-image capture followed by sending it to Gemini/ChatGPT, \cxr\ would be competitive; however, when live streaming assistance was available, \cxr\ ranked lowest. It is also worth mentioning that during our study, Gemini was free to use, whereas ChatGPT’s live multimodal features required a subscription with daily limits.

Overall, we regard the usability of \cxr{} as acceptable for an on-device AR prototype. 
First, the mean \gls{sus} of $60.0$ (SD $=17.1$; 95\% CI $[50.35,\,69.65]$) \emph{encompasses} the commonly cited benchmark of 68 for ``average'' usability~\cite{Lewis2018}, so average usability cannot be ruled out. 
Second, scores in the low 60s fall within the ``marginal/OK'' band on adjective/curved grading interpretations~\cite{Bangor2009}. 
Third, our result is comparable to \emph{GazePointAR} (62.1, SD $=20.0$) reported in a similar AR setting~\cite{Lee2024}, suggesting such values are typical for early head-worn AR interfaces. 
Moreover, relative to purely XR-native baselines (rather than state-of-the-art cloud assistants), we expect \cxr{} to compare more favorably due to explicit ROI selection, network-independent latency, and fully local operation.

\section{Conclusion}
\label{sec:conclusion}
We presented \cxr, a fully on-device multimodal \gls{vlm} system for XR that lets users select real-world objects and query them in natural language. \cxr\ provides a generic interface to deploy any suitably sized, ONNX-exportable VLM on XR headsets, and uses a segmentation-free, gaze-locked clipping window (GCW) for precise region-of-interest selection. This design reduces latency by avoiding a separate segmentation stage and mitigates pronoun ambiguity~\cite{tyler1977line} by ensuring the model receives exactly the user-selected crop. Because all processing occurs locally, \cxr\ preserves privacy and removes reliance on subscription-based or network-dependent cloud services. 

Empirically, \cxr\ delivers practical performance for interactive use: mean per-image inference times of \mbox{5.36--5.48\,s} across our two datasets, while remaining comparable to recent XR systems that depend on cloud-based AI (e.g., GazePointAR). GazePointAR reports a multi-stage pipeline with an overall latency of approximately 7.51\,s (including image capture, segmentation, and cloud-based inference)~\cite{Lee2024}.   Although \cxr\ does not yet match the absolute capability of state-of-the-art cloud assistants, it offers a viable, privacy-preserving alternative and a foundation for application-specific fine-tuning on XR data. We expect \cxr\ to serve as a baseline for XR applications that require natural-language interaction, strict data locality, or operation in connectivity-constrained environments, thereby contributing a practical bridge between advances in VLMs and real-world XR use.

\textbf{Future work.}
We plan to further improve efficiency through GPU-backed inference on ML2, model compression and quantization, and refined deployment strategies. On the modeling side, we aim to incorporate instruction-tuned checkpoints and task-specific fine-tuning for XR scenarios. We also intend to expand our usability studies to more complex tasks and real-world deployments, and to explore hybrid (privacy-preserving) client–server variants for larger models when appropriate.
 
\section*{Acknowledgments}
This research has been funded by KAUST Competitive Research Grants ORFS-CRG12-2024-6422. We thank Prof. Kiyoshi Kiyokawa from NAIST, Japan, for valuable discussions. We also thank Deng Luo and Da Li from our team at KAUST for their support and valuable input across various modules.


 
%

 \bibliographystyle{IEEEtran}  

\bibliography{Llamaxr}


 



\vspace{-33pt} 

\vfill
\end{document}